\documentclass[runningheads]{llncs}
\usepackage{graphicx}
\usepackage{amsmath}
\usepackage{amssymb}
\usepackage{amsfonts}
\usepackage{multirow}
\usepackage{color}
\usepackage{bm}
\usepackage{bbm}
\usepackage{mathtools} 
\usepackage[table]{xcolor}
\usepackage{arydshln}
\usepackage{subcaption}
\usepackage{booktabs}
\usepackage{colortbl}
\usepackage{graphicx}
\usepackage[table]{xcolor}
\usepackage{color}

\begin{document}

\title{Domain Adaptation \\ for Ulcerative Colitis Severity Estimation \\ Using Patient-Level Diagnoses}
\titlerunning{Weakly Supervised Domain Adaptation Leveraging Patient-Level Diagnoses}

\author{Takamasa Yamaguchi\inst{1} \and
Brian Kenji Iwana\inst{1} \and
Ryoma Bise\inst{1} \and
Shota Harada\inst{1} \and
Takumi Okuo\inst{1} \and
Kiyohito Tanaka\inst{2} \and
Kaito Shiku\inst{1}}
\authorrunning{T. Yamaguchi et al.}

\institute{Kyushu University, Japan \and
Kyoto Second Red Cross Hospital, Japan\\
    \email{takamasa.yamaguchi@human.ait.kyushu-u.ac.jp}}
\maketitle       
\begin{abstract}
The development of methods to estimate the severity of Ulcerative Colitis (UC) is of significant importance. However, these methods often suffer from domain shifts caused by differences in imaging devices and clinical settings across hospitals.
Although several domain adaptation methods have been proposed to address domain shift, they still struggle with the lack of supervision in the target domain or the high cost of annotation.
To overcome these challenges, we propose a novel Weakly Supervised Domain Adaptation method that leverages patient-level diagnostic results, which are routinely recorded in UC diagnosis, as weak supervision in the target domain.
The proposed method aligns class-wise distributions across domains using Shared Aggregation Tokens and a Max-Severity Triplet Loss, which leverages the characteristic that patient-level diagnoses are determined by the most severe region within each patient.
Experimental results demonstrate that our method outperforms comparative DA approaches, improving UC severity estimation in a domain-shifted setting.


\keywords{Ulcerative colitis \and Weakly supervised domain adaptation}
\end{abstract}
\section{Introduction}

\begin{figure}[t]
\centering
\includegraphics[width=1\textwidth]{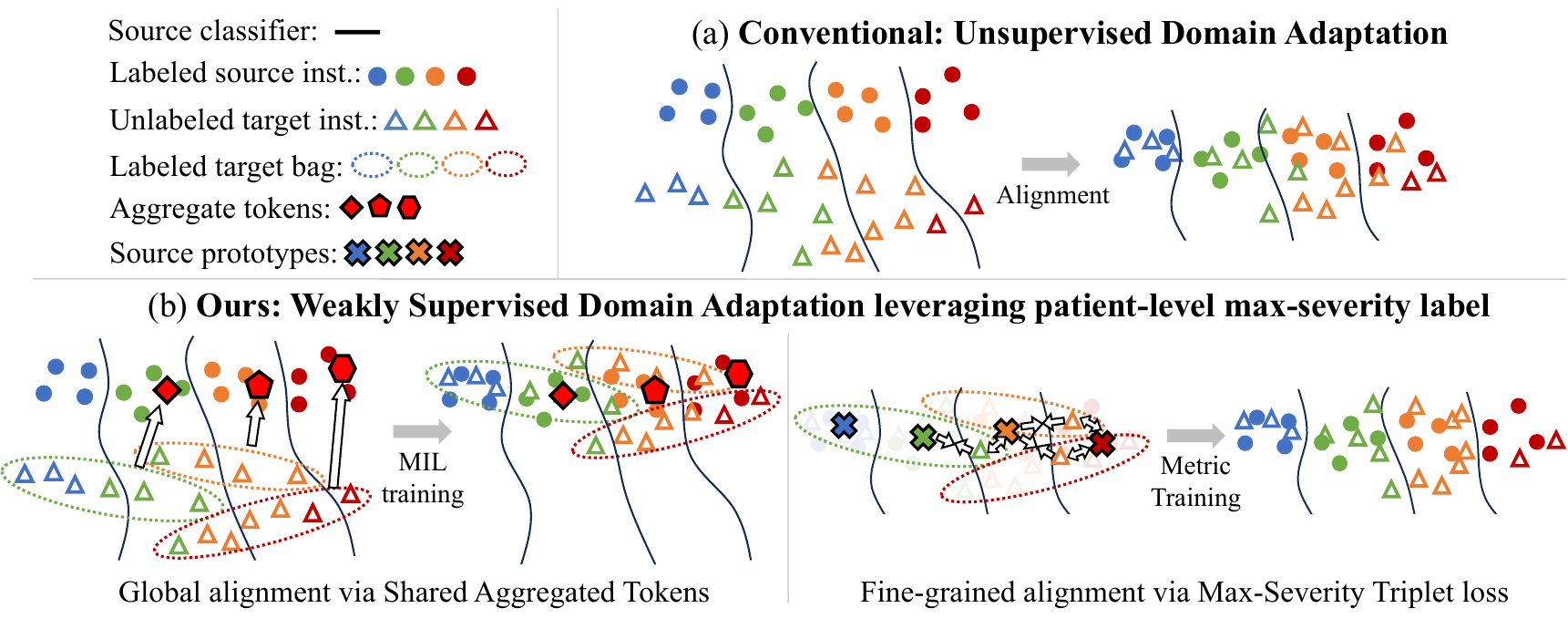}
\caption{(a) Conventional Unsupervised Domain Adaptation, which fails to achieve class-wise alignment across domains due to the lack of supervisory information in the target domain.
(b) The proposed Weakly Supervised Domain Adaptation, which leverages patient-level max-severity labels, achieves class-wise alignment through the Shared Aggregation Tokens strategy and the Max-Severity Triplet Loss.} \label{fig:ordinal}
\end{figure}


The diagnosis of Ulcerative Colitis (UC) is crucial for determining the appropriate clinical treatment. In the current UC diagnostic protocol, a patient undergoes endoscopic imaging of the colon, with approximately 20 to 40 images captured from different locations. The severity of UC, where a higher score indicates greater severity, is recorded in the patient's diagnosis report based solely on the image showing the most severe lesion, even though all images are reviewed during the diagnosis process~\cite{raimundo2016use,takabayashi2022development}. Importantly, the severity of individual images is not recorded.
However, there is a growing recognition of the importance of visualizing how the severity is distributed across different regions of the colon rather than focusing solely on the most severe areas~\cite{takabayashi2022development}. As such, there is a pressing need for automated methods to assess the severity of individual images taken during the diagnosis and to visualize the severity distribution across the entire colon.

Various approaches for automated severity assessment at both the image and patient levels have been explored.
For patient-level severity estimation, ordinal classification techniques, such as the K-rank algorithm, have been incorporated into Multi-Instance Learning (MIL)~\cite{schwab2022automatic,Shiku2025MIL}. However, these methods cannot estimate the image-level severity.
For image-level severity classification, many methods have been proposed to handle ordinal classes~\cite{kadota2022automatic,kadota2022deep,stidham2019performance,takenaka2020development,isbi_takezaki,polat2022class}.
For example, Kadota {\it et al.} \cite{kadota2022automatic} proposed an image-level severity estimation method that jointly learns regression and ranking tasks. 
These methods assume that the test dataset has the same distribution as the source data.

However, these methods encounter a significant challenge: domain shift. Domain shift occurs when there are discrepancies between the data distributions of the source (training) domain and the target (test) domain, often due to variations in imaging devices or hospital conditions. This leads to a degradation in performance when models trained on source domain data are applied to target domain data.

To address domain shift, Unsupervised Domain Adaptation (UDA)~\cite{ADDA,DANN} has been extensively studied. UDA assumes that no labeled data is available in the target domain and aims to align the overall data distributions between the source and target domains using only labeled data from the source domain. However, as shown in Fig.~\ref{fig:ordinal} (a), since UDA focuses only on aligning overall distributions, it struggles to align class-wise distributions within the domains.
 An alternative approach, Semi-Supervised Domain Adaptation (SSDA)~\cite{MME,CDAC,SSSD}, utilizes a small number of labeled target domain samples to align class-wise distributions. While SSDA improves performance over UDA, it still requires additional annotations, which often entail high costs due to the need for expert medical knowledge.

In this paper, we propose a novel Weakly Supervised Domain Adaptation problem that leverages routinely recorded patient-level diagnostic information in real clinical settings as weak supervision in the target domain.
As described above, the patient-level label, which represents the most severe score (the `max-severity label') among the images captured for each patient, is routinely stored as part of diagnostic records and can be utilized without requiring any additional annotation.
To the best of our knowledge, domain adaptation that leverages patient-level diagnosis has not been explored in the context of Ulcerative Colitis severity estimation.

To fully leverage patient-level max-severity labels, which are available only at the image set level, and to align class-wise feature distributions across domains, we propose a novel method that performs global distribution alignment via the Shared Aggregation Tokens Strategy and fine-grained alignment via the Max-Severity Triplet Loss.
The Shared Aggregation Tokens Strategy utilizes aggregation tokens, originally used for patient-level severity estimation in MIL~\cite{Shiku2025MIL}, for domain alignment.
As shown on the left of Fig.~\ref{fig:ordinal}(b), the aggregation tokens—trained in the source domain to predict max-labels and capture image-level class distributions—are frozen and reused during the training of max-label prediction in the target domain, thereby encouraging the class distribution in the target domain to approach that of the source domain and achieve coarse alignment.
The Max-Severity Triplet Loss, as illustrated  on the right Fig.~\ref{fig:ordinal} (b), leverages the characteristic of the max-label—that no images with a severity level higher than the max-label exist within a bag—and penalizes images in the target domain that are mistakenly predicted as a class more severe than the max-label, encouraging their distribution to align with the corresponding class in the source domain.

In experiments using two endoscopic image datasets, we confirmed that the proposed method outperforms conventional domain adaptation methods by leveraging patient-level diagnostic results.
Furthermore, the proposed method outperformed semi-supervised methods, even those using additional annotations on 5$\%$ of the dataset, despite requiring no extra annotations.

\section{Weakly Supervised Domain Adaptation Leveraging Patient-Level Diagnoses}

\subsection{Problem Setting}
Weakly Supervised Domain Adaptation leveraging patient-level diagnoses is formulated as domain adaptation on ordinal Multi-Instance Learning (MIL), where two domain datasets are provided as training data.
The source data, denoted as $\mathcal{D}^\mathrm{s} = \{\mathcal{B}_i^\mathrm{s}, Y_i^s \}$, is given with labels at both the instance ({\it i.e.,} image) and bag ({\it i.e.,} patient) levels. 
Each bag $\mathcal{B}_i^\mathrm{s} = \{\bm{x}_{i,j}^s, y_{i,j}^\mathrm{s} \}_{j=1}^{|\mathcal{B}_i^\mathrm{s}|}$ is defined as a set of $|\mathcal{B}_i^\mathrm{s}|$-th instances $\bm{x}_{i,j}^\mathrm{s}$ and instances labels $y_{i,j}^\mathrm{s}$.
The target data, denoted as $\mathcal{D}^\mathrm{t} = \{\mathcal{B}^\mathrm{t}_i, Y^\mathrm{t}_i \}$ is provided with labels at the bag level, while the instance-level labels for each bag in the target data, $\mathcal{B}_i^\mathrm{t} = \{\bm{x}_{i,j}^\mathrm{t} \}_{j=1}^{|\mathcal{B}_i^\mathrm{t}|}$ are not provided.
The goal of this paper is to train a classification model that estimates instance-level labels $ \hat{y}_{i,j}^\mathrm{t}$  in the target domain using these training data.

In both domains, bag-level and instance-level labels are defined with the $K$-th class, where $Y_i, y_{i,j} \in \{1, 2,\ldots,K\}$ and each class has an ordinal relationship $K \succ K-1,...,\succ 1$.
The bag-level label $Y_i$ is defined by the class with the highest severity among all instances in the bag.



\subsection{Domain Adaptation on Ordinal Multi-Instance Learning}

\begin{figure}[t]
\centering
\includegraphics[width=1\textwidth]{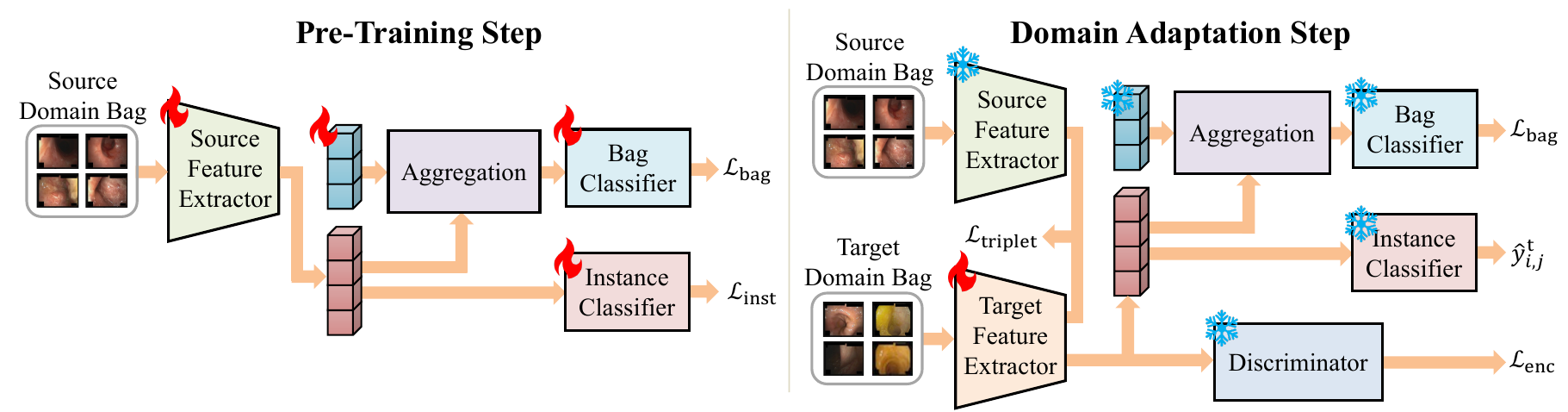}
\caption{Overview of proposed method} \label{fig:method}
\end{figure}

The proposed method is trained in two steps, as shown in Fig.~\ref{fig:method}. 
The first step involves pre-training the source feature extractor, aggregation token, and source classifiers using the source instance-level and bag-level labels.
In the second step, the target feature extractor is trained to align the domain-wise and class-wise distributions of the target data with those of the source data through global distribution alignment via the Shared Aggregation Tokens and fine-grained distribution alignment via the Max-Severity Triplet Loss.




\noindent
{\bf Pre-Training Step.}
This step aims to train the source feature extractor $g^\mathrm{s}\mathrm{inst}$, the source instance and bag classifiers $f^\mathrm{s}\mathrm{inst}$ and $f^\mathrm{s}_\mathrm{bag}$, as well as the aggregation tokens $\mathcal{T}={\bm{a}_1,\ldots,\bm{a}_{K-1}}$ through instance-level and bag-level classification training.

Image-level classification is conducted with instance-level labels using a standard supervised ordinal classification approach, such as the k-rank method~\cite{cao2020rank,niu2016ordinal}.
In bag-level classification training, inspired by the ordinal MIL method for patient-level severity estimation proposed by~\cite{Shiku2025MIL}, classification is performed on bag representations obtained using the aggregation tokens $\mathcal{T}$.
The aggregation tokens $\mathcal{T}$ compute attention scores for instances within a bag by measuring the similarity between themselves and the instance features, and generate the bag-level representation through weighted aggregation.
Each aggregation token is designed to aggregate the features of instances corresponding to a specific severity level. The $k$-th aggregation token, $\bm{a}_k$, is trained to assign high attention to instances of severity level $k$.
Through this training, each aggregation token is aligned with the class distribution in the source domain.

\noindent
{\bf Cross-Domain Global Distribution Alignment.}
The purpose of this step is to train the target feature extractor so that the domain-wise and class-wise data distributions are roughly aligned between the source and target domains.
During this step, the source feature extractor $g^\mathrm{s}_\mathrm{inst}$ and the classifiers $f^\mathrm{s}_\mathrm{inst}$ and $f^\mathrm{s}_\mathrm{bag}$ are frozen and only the target feature extractor $g^\mathrm{t}_\mathrm{inst}$ is trained.

First, to align the domain-wise distribution, adversarial learning is employed.
A domain discriminator $d(\cdot)$ is used to classify whether the data comes from the source or the target domain: $\mathcal{L}_\mathrm{disc}=-\sum_{j=1}^{|\mathcal{B}^\mathrm{s}_i|}\mathrm{log}(d(\bm{e}_{i,j}^\mathrm{s}))- \sum_{j=1}^{|\mathcal{B}^\mathrm{t}_i|} (\mathrm{log}(1-d(\bm{e}_{i,j}^\mathrm{t})))$.
Here, $\bm{e}_{i,j}^s$ and $\bm{e}_{i,j}^t$ denote the instance feature vectors in the source and target domains, respectively, obtained by the corresponding instance-level feature extractors. Specifically, $\bm{e}_{i,j}^s = g^\mathrm{s}_\mathrm{inst}(\bm{x}_{i,j}^\mathrm{s})$ and $\bm{e}_{i,j}^\mathrm{t} = g^\mathrm{t}_\mathrm{inst}(\bm{x}_{i,j}^\mathrm{t})$.
Next, the target feature extractor is trained so that it is mistaken for the source using an adversarial encoder loss: $\mathcal{L}_\mathrm{enc}=-\sum_{j=1}^{|\mathcal{B}^\mathrm{t}_i|}\mathrm{log}(d(\bm{e}_{i,j}^\mathrm{t}))$.

While adversarial learning can roughly align the domain-level distribution, aligning distributions at the class level remains challenging. 
To achieve class-wise alignment of instances across domains using only bag-level labels in the target domain, we propose a Shared Aggregation Token strategy that aims to align instance features in the target domain based on aggregation tokens pre-trained to capture the class distribution of instances in the source domain.
Specifically, when training bag classification in the target domain, the parameters of the aggregation tokens $\mathcal{T}$ pre-trained on the source domain are frozen, and only the feature extractor for the target domain $g^\mathrm{t}_\mathrm{inst}$ is updated. 
As a result, in order to correctly classify bags in the target domain, the instance features must align with the class-wise distribution of the source domain so that the fixed aggregation tokens can compute accurate attention scores for the target instances.
The Shared Aggregation Token strategy is optimized through the bag-level classification loss $\mathcal{L}_{\text{bag}}$ on the target domain.





\noindent
{\bf Fine-Grained Distribution Alignment with Max-Severity Triplet Loss.}
The purpose of this module is to align class distributions in detail across domains; however, since instance labels are not available in the target domain, achieving such alignment for each class is challenging.
To address this issue, a key idea of this module is to align the instance-level class distributions by leveraging the property of the max-severity labels that only instances with labels not exceeding the max label exist within a bag.
To do this, we propose the Max-Severity Triplet Loss, which consists of three components: an anchor, a positive, and a negative.
For selecting a Positive and Negative, we construct prototypes $\{\bm{p}_k^\mathrm{s}\}_{k=1}^{K}$ for each class using source data that are annotated with instance-level severity labels. Each prototype is computed as the average of instance features belonging to class $k$: 
$\bm{p}_k^\mathrm{s}=\frac{1}{J_k}\sum_{j=1}^{J_k} \bm{e}^{\mathrm{s}}_{k,j}$,
where $J_k$ is the number of source instances in class $k$ and $\bm{e}^{\mathrm{s}}_{k,j}$ are the instance features of class $k$.

Instances in the target domain that are predicted to have a severity higher than the bag label ${Y}^\mathrm{t}_{i}$ are considered to be data that do not align with the class distribution of the source domain. 
So, we use this as the Anchor and apply triplet loss. 
In this process, the Positive and Negative are selected as the source prototype corresponding to the same severity as ${y}^\mathrm{s}_{i,j}$ and a higher severity than ${y}^\mathrm{s}_{i,j}$, respectively.
Triplet loss $\mathcal{L}_\mathrm{triplet}$~\eqref{eq:L_Triplet} is imposed that ensures the distance between the Anchor instances and the Positive $\mathbf{p}^\mathrm{s}_+$ is smaller than the distance to the Negative $\mathbf{p}^\mathrm{s}_-$, or:
\begin{gather}
    \mathcal{L}_\mathrm{triplet}(\mathcal{B}^\mathrm{t}_i , \bm{p}^\mathrm{s}_+, \bm{p}^\mathrm{s}_-)= \sum_{j=1}^{|\mathcal{B}^\mathrm{t}_i|}\max(||\bm{e}_{i,j}^\mathrm{t}-\bm{p}^\mathrm{s}_+||-||\bm{e}_{i,j}^\mathrm{t}-\bm{p}^\mathrm{s}_-||+\xi,0)\label{eq:L_Triplet},\\
    \mathrm{s.t.}\ {Y}^\mathrm{t}_i\leq{K-1}, {Y}^\mathrm{t}_i < \hat{y}_{i,j}^\mathrm{t},
\end{gather}
where $||\cdot||$ denotes the Euclidean distance and $\xi$ is the margin. 


Then, the total loss is calculated with a hyperparameter $\alpha$ controlling the weight of the triplet loss, as follows:
\begin{align}
\label{eq:total}
\mathcal{L}_\mathrm{target}=\mathcal{L}_\mathrm{bag}+\mathcal{L}_\mathrm{enc}+\alpha\mathcal{L}_\mathrm{triplet}.
\end{align}


\section{Experimental Results}

\noindent
{\bf Datasets.\ }
To validate the effectiveness of our proposed method, we used two datasets ({\bf LIMUC})~\cite{polat2023improving} and a private dataset ({\bf Private}) as different domains.
{\bf LIMUC} consists of 11,276 images collected from 564 patients, with the number of images in the bag from 1 to 105, and an average of 20 images per bag.
This dataset has image-level labels and allows for obtaining a patient-level max-severity label based on the most severe region for each patient.
{\bf Private} collected from an anonymous hospital, it has patient-specific max-severity labels recorded during routine diagnoses.
Here, in this experiment, additional annotations were performed on an image basis for the purpose of evaluation.
Each image-level is annotated with a diagnostic label representing severity-level ranging from severity 0 to severity 3.
In both datasets, as the target domain, only patient-level labels are used for training, excluding image-level labels.


\noindent
{\bf Implementation Detail.\ }
For the feature extractor, we used ResNet18~\cite{he2016deep} pre-trained on the ImageNet dataset~\cite{deng2009imagenet}. 
The proposed network was optimized using the Adam optimizer~\cite{adam}. 
In the Pre-Training step, the learning rate for the feature extractor and the instance classifiers are 3e-6, and bag classifier is 1e-5, respectively, for 1,500 epochs with a mini-batch size of 16, and early stopping of 100 patients.
To address class imbalance, 
in each step oversampling based on the number of instance-level and bag-labels is used.
In the Domain Adaptation Step, the learning rates for the discriminator and feature extractor are 1e-4 and 1e-6 respectively, for a fixed 150 epochs with a mini-batch size of 16, and $\alpha$ in the $\mathcal{L}_\mathrm{target}$ of 0.01. 

\noindent
{\bf Evaluation Metrics.\ }
To evaluate the performance of our proposed method, we employed three metrics: classification accuracy, Macro-F1 which is robust to data imbalance, and Quadratic Weighted Kappa (Kappa)~\cite{polat2022class} which takes ordinal relationships into account. 
5-fold cross-validation was used for the evaluations.

\noindent
{\bf Comparative Evaluation.\ }
We compare the proposed method with several domain adaptation methods, including two UDA methods, ADDA~\cite{ADDA} and DANN~\cite{DANN}, and three SSDA methods, MME~\cite{MME}, CDAC~\cite{CDAC}, and 
S$^3$D~\cite{SSSD}.
The UDA methods align the overall data distributions through adversarial learning without target labels and the SSDA methods utilize some labeled data in the target domain.
It should be noted that SSDA methods use additional supervised labels in the target domain, while the proposed method trains without annotation costs by leveraging routinely recorded patient-level diagnostic records. 
In this experiment, the SSDA method uses 1\%, 3\%, and 5\% of the images in each class as labeled training data. 


\begin{table}[t]
    \def\@captype{table}
      \makeatother
        \centering
        \caption{Comparison with UDA and SSDA methods. `Target Label' indicates the type of label in the target domain. `Instance Label Ratio' indicates the ratio of additional annotated labels. The best results are in {\bf bold}.}  
        \scalebox{0.80}{
        \begin{tabular}{c|c||c||ccc||ccc} \hline
        &&&\multicolumn{3}{c||}{LIMUC to Private} & \multicolumn{3}{c}{Private to LIMUC}\\
        Target Label & Instance Label Ratio & Method & Accuracy & Kappa & Macro-F1 & Accuracy & Kappa & Macro-F1 \\ \hline \hline
                \multirow{2}{*}{Unsupervised} 
        &\multirow{2}{*}{0\%}& ADDA  & 0.521& 0.575& 0.364& 0.604& 0.645& 0.529\\
        && DANN & 0.429& 0.500& 0.352& 0.560& 0.446& 0.423\\ 
        \hline

        \multirow{9}{*}{Semi supervised}
        & \multirow{3}{*}{1\%}  & MME & 0.610 & 0.593 & 0.469 & 0.670 & 0.695 & 0.541\\
        &                       & CDAC & 0.588 & 0.545 & 0.446 & 0.593 & 0.571 & 0.506\\
        &                       & S$^{3}$D & 0.653 & 0.641 & 0.507 & 0.666 & 0.684 & 0.522\\
        \cdashline{2-9}
        & \multirow{3}{*}{3\%}  & MME & 0.588 & 0.550 & 0.476 & 0.679 & 0.727 & 0.570\\
        &                       & CDAC & 0.670 & 0.643 & 0.512 & 0.607 & 0.607 & 0.532\\
        &                       & S$^{3}$D & 0.674 & 0.655 & 0.537 & 0.677 & 0.726 & 0.567\\
        \cdashline{2-9}
        & \multirow{3}{*}{5\%} & MME & 0.608& 0.593 & 0.496 & 0.697 & 0.753 & \bf{0.595}\\
        &                      & CDAC & 0.651 & 0.628 & 0.522 & 0.633 & 0.655 & 0.562\\
        &                      & S$^{3}$D & 0.668 & 0.652 & 0.534 & 0.692 & 0.738 & 0.586\\
        \hline
        \rowcolor{gray!10} Patient-level labels&0\%& Ours & \bf{0.714}& \bf{0.746}& \bf{0.603}& \bf{0.706}& \bf{0.787}& 0.594\\ %

        \hline        
        \end{tabular}
        }
        \label{tab:comparison}
\end{table}

Table~\ref{tab:comparison} shows the comparison results with the comparative methods.
Due to the lack of class information in the target domain, ADDA and DANN struggle to align class distributions across domains.
Therefore, they were unable to achieve sufficient performance.
MME, CDAC, and S$^3$D achieved higher performance than the UDA methods due to the additional instance information in the target domain.
However, 5\% of the entire training dataset corresponds to approximately 400 images, requiring extremely high annotation costs by medical experts.
In contrast, the proposed method utilizes routinely recorded patient-level diagnostic records, achieving the highest performance among all methods in the LIMUC to Private experiment. Additionally, in the Private to LIMUC experiment, it attains the highest accuracy and kappa scores, while also achieving Macro-F1 performance comparable to the best-performing method that uses 5\% labeled data.
These results suggest that the proposed method can utilize patient-level diagnostic records, which do not require annotation costs, as weak labels and align class-wise distributions across domains.

\begin{table}[tb]
    \def\@captype{table}
      \makeatother
        \centering
        \caption{Effectiveness of each module. The best results are in {\bf bold}.}    
        \scalebox{0.8}{
     \begin{tabular}{c||c|c|c|| ccc ||ccc}  \hline
     \multirow{2}{*}{Method} &  \multirow{2}{*}{Adv.} & \multirow{2}{*}{Agg. Token} & \multirow{2}{*}{Triplet} &   \multicolumn{3}{c||}{LIMUC $\rightarrow$ Private} & \multicolumn{3}{c}{Private $\rightarrow$ LIMUC} \\
        & &&&  Accuracy & Kappa & Macro-F1 & Accuracy & Kappa & Macro-F1\\ \hline \hline
        \rowcolor{gray!15} Ours &$\checkmark$&$\checkmark$&$\checkmark$& \bf{0.714}& \bf{0.746}& \bf{0.603}& \bf{0.706}& \bf{0.787}& 0.594\\
        w/o Triplet &$\checkmark$&$\checkmark$&& 0.658& 0.703& 0.576& 0.705& \bf{0.787}& \bf{0.605}\\
        w/o Triplet\&AT&$\checkmark$&&& 0.521& 0.574& 0.363& 0.604& 0.645& 0.529\\
        w/o Adv\&Triplet\&AT &&&& 0.237& 0.300& 0.212 &0.580& 0.397& 0.344\\
        \hline        
        \end{tabular}
        }
        \label{tab:ablation}
\end{table}

\noindent
{\bf Effectiveness of Each Module.\ }
We conduct an ablation study to evaluate the effectiveness of the Shared Aggregation Tokens and Max-Severity Triplet Loss by comparing our method with three ablation methods.
Table~\ref{tab:ablation} shows the results of the ablation study. 
We compare the proposed method to models without the triplet loss (w/o Triplet), without the triplet loss and the shared aggregation token (w/o Triplet\&AT), and a method trained without triplet loss, the shared aggregation token and adversarial learning  (w/o Adv\&Triplet\&AT), i.e. trained only on the source domain and tested on the target domain.
The results show that although adversarial learning improves performance, it is not sufficient on its own. By additionally introducing the proposed Shared Aggregation Token and Max-Severity Loss to achieve class-wise distribution alignment across domains, performance is significantly enhanced.



\begin{figure}[t]
\centering
\includegraphics[width=0.60\textwidth]{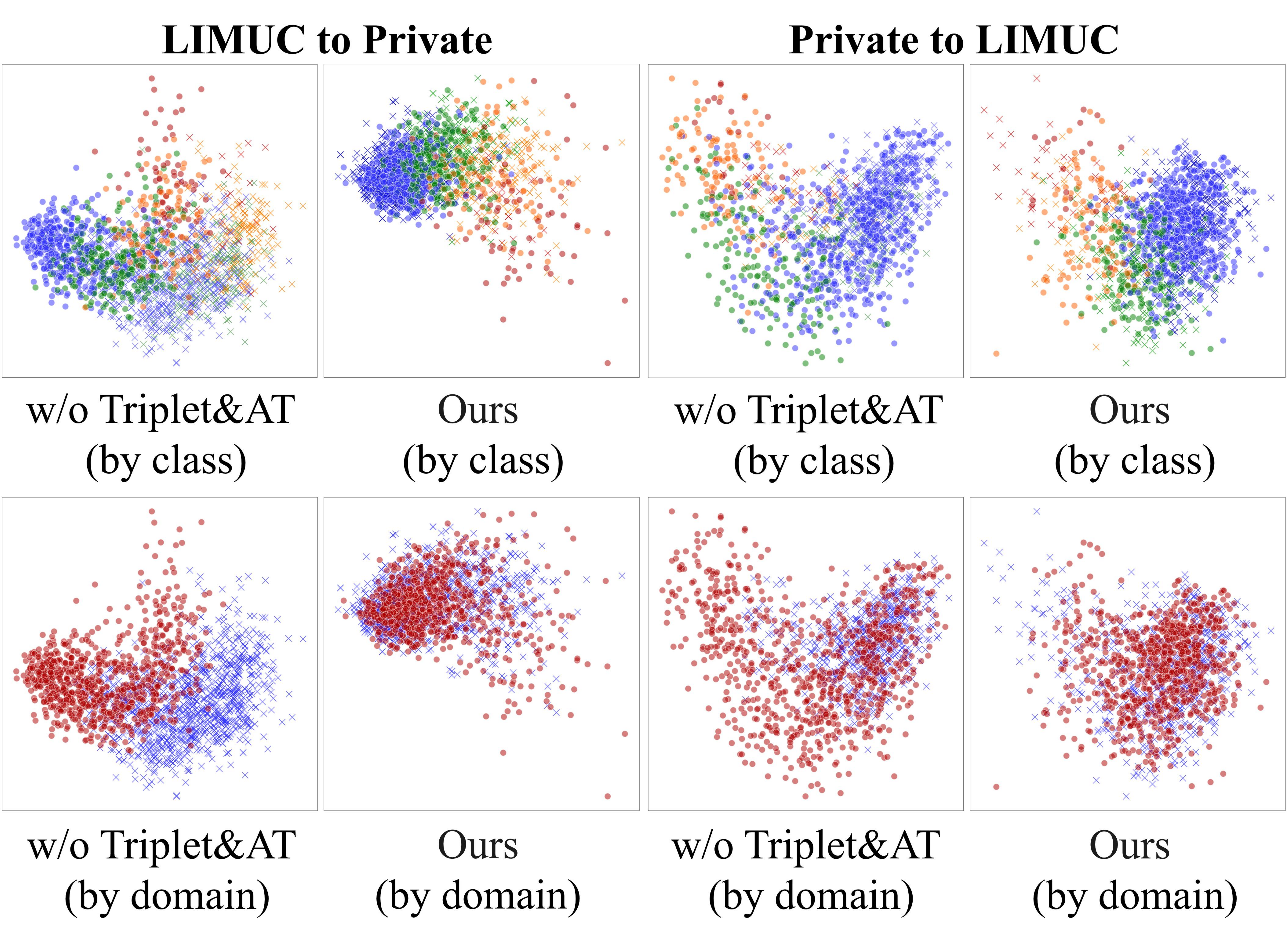}
\caption{A visualization of the feature space using PCA of the target domain before and after the proposed method. In the upper row, the classes, severity 0, 1, 2, and 3, are shown in blue, green, orange, and red, respectively. In the lower row, the source domain is shown in blue and the target domain in red.} \label{fig:featurespace}      
\end{figure}

To demonstrate that the proposed method aligns feature distributions across domains at the class level, Fig.~\ref{fig:featurespace} visualizes the data distributions using Principal Component Analysis (PCA).
These results indicate that the model trained with adversarial learning alone (w/o Triplet \& AT) fails to effectively align class-wise distributions across domains, whereas the proposed method, incorporating the Shared Aggregation Token and Max-Severity Loss, achieves improved class-wise distribution alignment.


\section{Conclusion}
In this study, we proposed a Weakly Supervised Domain Adaptation method for UC severity estimation that leverages routinely recorded patient-level diagnostic results as weak supervisory labels in the target domain.
To align class-wise distributions across domains using patient-level max-severity diagnoses, we proposed a novel method consisting of Shared Aggregation Tokens and a Max-Severity Triplet Loss.
The experimental results confirmed the effectiveness of the proposed method. Furthermore, a comparison with semi-supervised methods that require additional annotations demonstrated that our method achieves high performance without incurring additional annotation costs.

\noindent
{\bf Acknoledgement}:This work was partially supported by KAKEN JP23K16949, JP2318509, SIP JPJ012425, and ASPIRE JPMJAP2403 .

%
%


\bibliographystyle{splncs04}
\bibliography{mybibliography}

\end{document}